\documentclass{article}

\usepackage[preprint]{neurips_2025}

\usepackage[utf8]{inputenc} 
\usepackage[T1]{fontenc}    
\usepackage{hyperref}       
\usepackage{url}            
\usepackage{booktabs}       
\usepackage{amsfonts}       
\usepackage{nicefrac}       
\usepackage{microtype}      

\usepackage{latexsym}
\usepackage{wrapfig}
\usepackage{arydshln}
\usepackage{multirow} 
\usepackage{hyperref}
\usepackage{booktabs}  
\usepackage{amsthm}
\usepackage{enumerate}
\usepackage{wrapfig}
\usepackage[table]{xcolor}
\usepackage{colortbl}
\usepackage{multirow} 

\usepackage{amsmath,amsfonts}
\usepackage{algorithmic}
\usepackage{algorithm}
\usepackage{array}
\usepackage[caption=false,font=normalsize,labelfont=sf,textfont=sf]{subfig}
\usepackage{textcomp}
\usepackage{stfloats}
\usepackage{url}
\usepackage{verbatim}
\usepackage{graphicx}
\usepackage{cite}
\usepackage{wrapfig}

\title{Meta-Chunking: Learning Text Segmentation and Semantic Completion via Logical Perception}

\author{
  \,Jihao Zhao\textsuperscript{1} \quad
  Zhiyuan Ji\textsuperscript{1} \quad
  Yuchen Feng\textsuperscript{2} \quad
  Pengnian Qi\textsuperscript{2} \quad
  Simin Niu\textsuperscript{1} \quad
  Bo Tang\textsuperscript{2} \quad \\
  \,
  \textbf{Feiyu Xiong}\textsuperscript{\textbf{2}} \quad
  \textbf{Zhiyu Li}\textsuperscript{\textbf{2}}\thanks{Corresponding author: \texttt{lizy@iaar.ac.cn}} \\
  \,\textsuperscript{1}School of Information, Renmin University of China, Beijing, China \\
  \,\textsuperscript{2}Institute for Advanced Algorithms Research, Shanghai, China \\
}

\begin{document}

\maketitle

\begin{abstract}
While Retrieval-Augmented Generation (RAG) has emerged as a promising paradigm for boosting large language models (LLMs) in knowledge-intensive tasks, it often overlooks the crucial aspect of text chunking within its workflow. This paper proposes the Meta-Chunking framework, which specifically enhances chunking quality through a dual strategy that identifies optimal segmentation points and preserves global information. Initially, breaking limitations of similarity-based chunking, we design two adaptive chunking techniques based on uncertainty, namely Perplexity Chunking and Margin Sampling Chunking, by utilizing the logical perception capabilities of LLMs. Given the inherent complexity across different texts, we integrate meta-chunk with dynamic merging, striking a balance between fine-grained and coarse-grained text chunking. Furthermore, we establish the global information compensation mechanism, encompassing a two-stage hierarchical summary generation process and a three-stage text chunk rewriting procedure focused on missing reflection, refinement, and completion. These components collectively strengthen the semantic integrity and contextual coherence of chunks. Extensive experiments demonstrate that Meta-Chunking effectively addresses challenges of the chunking task within the RAG system, providing LLMs with more logically coherent text chunks. Additionally, our methodology validates the feasibility of implementing high-quality chunking tasks with smaller-scale models, thereby eliminating the reliance on robust instruction-following capabilities. Our code is available at \href{https://github.com/IAAR-Shanghai/Meta-Chunking}{https://github.com/IAAR-Shanghai/Meta-Chunking}.
\end{abstract}

\section{Introduction}
Retrieval-augmented generation (RAG), as a technical paradigm that integrates information retrieval with generative models, effectively mitigates inherent limitations of large language models (LLMs), such as data freshness \cite{he2022rethinking}, hallucinations \cite{chen2023hallucination,zuccon2023chatgpt,liang2024internal}, and the lack of domain-specific knowledge \cite{li2023chatgpt,shen2023chatgpt}. As the core architecture for knowledge-intensive tasks \cite{lazaridou2022internet}, its efficacy is fundamentally constrained by the optimization boundary of the synergistic retrieval-generation mechanism, because the quality of retrieved text chunks directly determines the performance ceiling \cite{li2022survey,tan2022tegtok,lin2023li}, and the foundation of this process lies in the text chunking. Optimally segmenting documents into semantically complete and coherent chunks not only enhances the generation accuracy of LLM by concentrating information and reducing redundancy \cite{xu2023berm,su2024dragin}, but also significantly improves the processing efficiency of the system while reducing computational resource consumption \cite{besta2024multi}.

As a preprocessing unit within the RAG system, this process is often overlooked and has consequently received insufficient in-depth investigation \cite{sidiropoulos2022analysing,zhuang2024efficientrag,kim2024adaptive}. Current mainstream methods primarily rely on rules or semantic similarity \cite{zhang2021sequence,langchain,lyu2024crud}. Although these approaches are engineering-friendly, they typically fail to capture the nuanced logical dependencies between sentences. As illustrated in Figures \ref{fig:rag_pipeline} and \ref{fig:luoji}, chunks with logical progression are often incorrectly segmented due to low cosine similarity, leading to retrieval results deviating from the core semantic unit. Recently proposed LumberChunker \cite{duarte2024lumberchunker}, while invoking LLM APIs to more accurately identify content divergence points, requires models with advanced instruction-following capabilities, thereby incurring substantial resource costs. This raises a practical question: How can we fully utilize the powerful reasoning capabilities of LLMs while efficiently accomplishing the text chunking task at a lower cost?

Inspired by these observations, this paper proposes the \textbf{Meta-Chunking} framework, which synergistically optimizes the logical perception capabilities of LLMs with information integrity constraints to specifically address the issue of logical discontinuities in text chunking. We design two uncertainty-based adaptive boundary detection algorithms: Perplexity (PPL) Chunking and Margin Sampling (MSP) Chunking. These algorithms leverage the implicit and explicit evaluation capabilities of LLMs for logical coherence, respectively, to identify chunk boundaries. Meanwhile, the resulting meta-chunks are treated as independent logical units, and a dynamic merging strategy is introduced to achieve a balance between fine-grained and coarse-grained segmentation. On the other hand, to further enhance the cognitive completeness of text chunks, we construct an information compensation pipeline: (1) Implementing a missing-aware rewriting mechanism during the post-chunking phase, which systematically repairs semantic discontinuities caused by segmentation through a three-stage optimization process of missing reflection, refinement, and completion. (2) Adopting a two-layer summarization technique for each text chunk, we extract core knowledge anchors from both the document-level macro themes and the paragraph-level micro semantics, thereby further improving the global recall rate of chunks.  It is noteworthy that due to the scarcity of relevant datasets in the chunking domain, we carefully prepare training data for aforementioned methods and fine-tune small language models (SLMs) to achieve efficient application. 

We summarize contributions of this work as follows:
\begin{itemize}
    \item Through lightweight chunking algorithm design, the logical analysis capability of LLMs is decoupled into computable the PPL features and MSP indicators, achieving identification of textual logical boundaries and  dynamic balance of chunking granularity.
    \item We establish a information compensation mechanism that collaboratively executes through a three-stage missing-aware rewriting process and a two-stage context-aware summary generation, repairing the semantic discontinuities in text chunks.
    \item To verify the effectiveness of our proposed Meta-Chunking framework, we conduct multidimensional experiments and analyses using five datasets. The results indicate that this framework delivers more logically coherent text chunks to the RAG system, demonstrating the feasibility of achieving high-quality chunking tasks on SLMs.
\end{itemize}

\section{Related Works}

\subsection{Text Chunking in RAG}
By expanding the input space of LLMs through introducing retrieved text chunks \cite{guu2020retrieval,lewis2020retrieval}, RAG significantly improves the performance of knowledge-intensive tasks \cite{ram2023context}. Text chunking plays a crucial role in RAG, as ineffective chunking strategies can lead to incomplete contexts or excessive irrelevant information, thereby hurting the performance of question answering (QA) systems \cite{yu2023chain}. Besides typical granularity levels like sentences or paragraphs \cite{lyu2024crud,gao2023retrieval}, there are other advanced methods available. \cite{chen2023dense} introduced a novel retrieval granularity called Proposition, which is the smallest text unit that conveys a single fact. This method excels in fact-based texts like Wikipedia. However, it may not perform ideally when dealing with content that relies on flow and contextual continuity, such as narrative texts, leading to the loss of critical information. Meanwhile, LumberChunker \cite{duarte2024lumberchunker} iteratively harnesses LLMs to identify potential segmentation points within a continuous sequence of textual content, showing some potential for LLMs chunking. However, this method demands a profound capability of LLMs to follow instructions and entails substantial consumption when employing the Gemini model.

\subsection{Uncertainty Theory of LLMs}
Quantifying uncertainty in LLMs is currently an active research direction in the field of artificial intelligence \cite{zhang2024vl,li2025language,da2025understanding}. Information theory provides a solid theoretical foundation and a suite of mathematical tools to measure the inherent degree of uncertainty in probability distributions or signals. For instance, Entropy is employed to gauge the randomness of a model's prediction for the next token \cite{atf2025challenge}. Semantic Entropy further extends this concept to encompass clusters of semantically similar generated sequences \cite{farquhar2024detecting}. Perplexity \cite{liu2025uncertainty}, a classic metric for evaluating LLMs, indirectly reflects the strength of logical relationships between sentences by measuring the model's surprise regarding sequential data. Additionally, Mutual Information is capable of quantifying the amount of information shared between different random variables, making it useful for assessing cognitive uncertainty among various outputs of different models \cite{abbasi2024believe}.

\section{Methodology}
\subsection{Text Chunking of Meta-Chunking}
Our approach is grounded in a core principle: allowing variability in chunk size to more effectively capture and maintain the logical integrity of content. This dynamic adjustment of granularity ensures that each segmented chunk contains a complete and independent expression of ideas, thereby avoiding breaks in the logical chain during the segmentation process. This not only enhances the relevance of document retrieval but also improves content clarity.

\begin{figure*}[t]
    \centering
    \includegraphics[width=\textwidth]{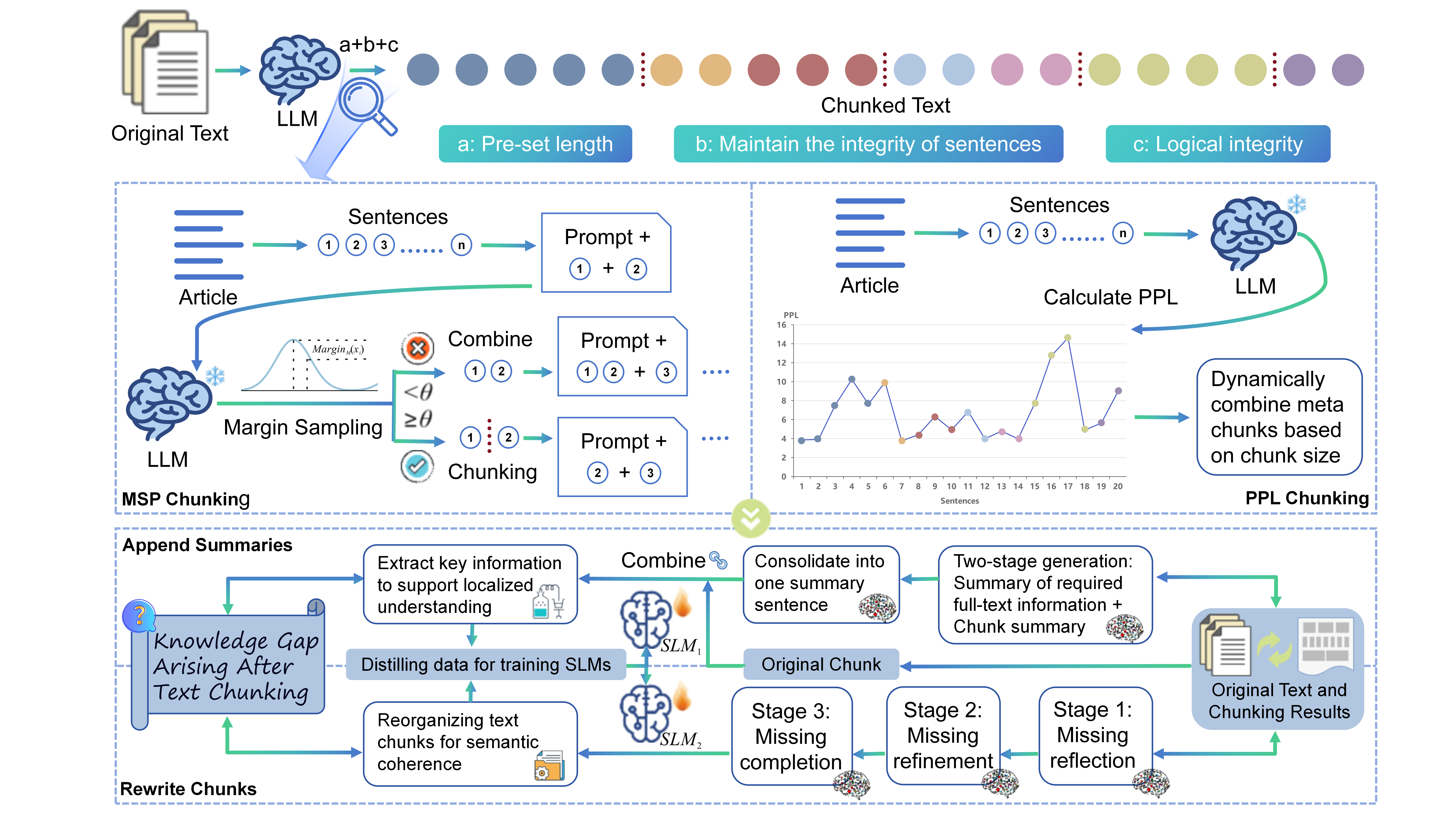}
    \caption{Overview of the entire process of Meta-Chunking. Each circle represents a complete sentence, and the sentence lengths are not consistent. The vertical lines indicate where to segment. Circles with the same background color represent a meta-chunk, which is dynamically combined to make the final chunk length meet user needs.}
    \label{fig:metachunker_pipeline}
\end{figure*}

As illustrated in Figure~\ref{fig:metachunker_pipeline}, our method integrates the advantages of traditional text segmentation strategies, such as adhering to preset chunk length constraints and ensuring sentence structural integrity, while enhancing the ability to guarantee logical coherence during the segmentation process. We refer to each text chunk obtained through segmentation as a \textbf{Meta-Chunk}, which consists of a collection of sequentially arranged sentences within a paragraph. These sentences not only have semantic relevance but, more importantly, also contain profound linguistic logical connections, including but not limited to general-specific, parallel, sequential, and illustrative relationships, as shown in Figure \ref{fig:luoji}. Through observation, it is found that there are often tight logical connections between consecutive sentences within a meta-chunk. However, these sentences exhibit low semantic similarity due to their divergent content representations. As mentioned in \cite{qu2024semantic}, semantic chunking has failed to demonstrate advantages across multiple experimental paradigms. We believe that this phenomenon is closely related to the original theoretical modeling intentions of semantic similarity algorithms. These methods essentially model the degree of semantic overlap between texts to quantify the correlation between two paragraphs or between a sentence and a paragraph. Nevertheless, sentences at the micro level that have logical associations but express different content limit their applicability. The detailed analysis is presented in Appendix \ref{appendix2}. In order to address the aforementioned issue, we implement the following strategies based on the uncertainty theory in LLMs.

\subsubsection{Perplexity Chunking}
Given a text, the initial step involves segmenting it into a collection of sentences denoted as $(x_{1}, x_{2},\dots,x_{n})$, with the ultimate goal being to further partition these sentences into several chunks, forming a new set $(X_{1},X_{2},\dots,X_{k})$, where each chunk comprises a coherent grouping of the original sentences. We split the text into sentences and use the model to calculate the PPL of each sentence $x_{i}$ based on the preceding sentences:
\begin{eqnarray}
\text{PPL}_{M}(x_{i})=\frac{ {\textstyle \sum_{k=1}^{K}}\text{PPL}_{M}(t_{k}^{i}|t_{<k }^{i},t_{<i}) }{K} 
\end{eqnarray}
where $K$ represents the total number of tokens in $x_{i}$, $t_{k}^{i}$ denotes the $k$-th token in $x_i$, and $t_{<i}$ signifies all tokens that precede $x_{i}$. To locate the key points of text segmentation, the algorithm further analyzes the distribution characteristics of $\text{PPL}_{seq}=(\text{PPL}_{M}(x_{1}),\text{PPL}_{M}(x_{2}),\dots,\text{PPL}_{M}(x_{n}))$, particularly focusing on identifying minima.

Our primary focus is on two types of minimum points: when the PPL on both sides of a point are higher than at that point, and the difference on at least one side exceeds the preset threshold $\theta$; or when the difference between the left point and the point is greater than $\theta$ and the right point equals the point value. These minima are regarded as potential chunk boundaries. If the text exceeds the processing range of LLMs or device, we strategically introduce a key-value (KV) caching mechanism. Specifically, the text is first divided into several parts according to tokens, forming multiple subsequences. As the PPL calculation progresses, when the GPU memory is about to exceed the server configuration or the maximum context length of LLMs, the algorithm appropriately removes KV pairs of previous partial text, thus not sacrificing too much contextual coherence.

\subsubsection{Margin Sampling Chunking}
It is noteworthy that LumberChunker \cite{duarte2024lumberchunker} encounters difficulties when applied to smaller models, primarily due to its requirement for generating text in a specified format and subsequent regular expression extraction. To address this limitation, we introduce the MSP strategy that analyzes the marginal probability distribution during model decision-making to determine whether chunking should be performed. The method can be formulated as:
\begin{eqnarray}
\text{Margin}_{M}(x_{i})=P_{M}\left(y=k_{1}|\text{Prompt}(x_{i},X^{'})\right )-P_{M}\left(y=k_{2}|\text{Prompt}(x_{i},X^{'})\right)
\end{eqnarray}
where $(k_{1},k_{2})$ indicates a binary decision between $yes$ or $no$ for a segmentation judgment. $\text{Prompt}(x_{i},X^{'})$ represents forming an instruction between $x_{i}\in \{x_{l}\}_{l=1}^{n}$ and $X^{'}$, regarding whether they should be merged, where $X^{'}$ encompasses either a single sentence or multiple sentences. Through the probability $P_{M}$ obtained by model $M$, we can derive the probability difference $\text{Margin}_{M}(x_{i})$ between the two options. Subsequently, by contrasting $\text{Margin}_{M}(x_{i})$ with the threshold $\theta$, a conclusion can be drawn regarding whether the two sentences should be segmented. Moreover, setting a threshold for decision criteria is a common requirement across all strategies, and we bring in a dynamic threshold mechanism.
Specifically, in the initialization phase of the $\theta$, we assign it a starting value of 0. Subsequently, we fine-tune $\theta$ by keeping track of historical $\text{Margin}_{M}(x_{i})$ values and computing their mean, thereby enabling more flexible adjustment of chunking.

\subsubsection{Dynamic Merging}
To address diverse chunking needs of users, merely adjusting the threshold to control chunk size sometimes leads to uneven chunking sizes as the threshold increases, as shown in Appendix \ref{appendix5}. Therefore, we propose a strategy combining meta-Chunk with dynamic merging, aiming to flexibly respond to varied chunking requirements. Firstly, we employ either PPL Chunking or MSP chunking to partition the document into a series of meta-chunks, denoted as $(c_{1},c_{2},\dots,c_{\alpha})$. Traditional chunking methods treat sentences as independent logical units, whereas we adopt meta-chunks as independent logical units. Subsequently, according to the user-specified chunk length $L$, we iteratively merge adjacent meta-chunks until the total length satisfies or approximates the requirement. Specifically, if $\text{len}(c_{1},c_{2},c_{3})=L$ or $\text{len}(c_{1},c_{2},c_{3})<L$ while $\text{len}(c_{1},c_{2},c_{3},c_{4})>L$, then $c_{1},c_{2},c_{3}$ are regarded as a complete chunk.

\subsection{Semantic Completion of Meta-Chunking}
\label{sec33z}
To address the semantic gap issue arising from the loss of contextual information in text chunking, we propose a globally enhanced text rewriting and summary generation mechanism. Specifically, we leverage a LLM as a discriminator to examine whether each chunk suffers from semantic deficiencies, and if so, initiate the rewriting process in Section \ref{sec33_1}. After handling these deficiencies, we perform the summary generation in Section \ref{sec33_2} on all chunks to further improve recall, laying a solid foundation for ultimately enhancing QA performance. The detailed design scheme is elaborated in Appendix \ref{appendix_add1}.

\subsubsection{Globally Augmented Text Chunk Rewriting}
\label{sec33_1}
\textbf{Preprocessing (Optional)}\quad For extremely long documents that present challenges for full ingestion by LLM, an inter-chunk relevance analysis leveraging semantic embeddings is employed. 
This process involves generating vector representations for each chunk using a semantic similarity model and quantifying the strength of their semantic associations by calculating the cosine similarity between these vectors. Such an approach facilitates the identification of potential contextual information pertinent to the current chunk.

\textbf{Stage 1 (Missing Reflection)}\quad Utilizing an LLM, and incorporating the potentially relevant information identified during the preprocessing phase, each chunk undergoes an in-depth reflective analysis. The core task is to explicitly identify which premises, backgrounds, related facts, or conclusive statements are missing from the current chunk. The LLM should comprehensively list the areas where information is missing and specify the information that needs to be supplemented.

\textbf{Stage 2 (Missing Refinement)}\quad This phase is dedicated to score and filter the potentially missing information detected in the previous stage. We aim to prevent the introduction of irrelevant or erroneous supplementary content, thereby ensuring the precision of the augmentation process.

\textbf{Stage 3 (Missing Completion)}\quad Based on the refined omission loci and the requisite information confirmed in the preceding stage, the LLM is prompted to integrate these informational segments with the current text chunk. The goal is to generate a new chunk that is contextually seamless, semantically natural, and effectively achieves robust inter-chunk information fusion.

\subsubsection{Context-Aware Summary Generation}
\label{sec33_2}
The primary objective of this part is to generate a concise summary, enriched with global information, for each text chunk, thereby further augmenting the contextual awareness of the chunk.
\begin{enumerate}[(1)]
\item The model utilizes global information to generate a supplementary summary for the target text chunk. This process is designed to compensate for the discourse background and external relational information that the chunk may lack due to segmentation.
\item With respect to the content of the chunk itself, the model independently generates a local summary that encapsulates its core viewpoint. Subsequently, the aforementioned two summaries are fused and refined into an enhanced summary sentence that can articulate the content of the chunk from a global perspective.
\end{enumerate}

To support the proposed rewriting and summary generation components, we construct 20,000 training data samples for each of them, adhering to the process described above. Meanwhile, we opt for full fine-tuning of the SLM. For an input sequence $X$ and a target output sequence $Y = (y_1, y_2, ..., y_T)$, the loss function is defined as:

\begin{equation} 
L(\theta) = -\frac{1}{N} \sum_{t = 1}^{T} \log P(y_t | y_{<t}, X; \theta) \label{eq:cross_entropy_loss} 
\end{equation} 
where $P(y_t | y_{<t}, X; \theta)$ represents the probability that the model predicts the true target token $y_t$ given the input $X$ and the previously generated prefix $y_{<t}$, $\theta$ denotes the model parameters, and $N$ is the number of samples in a batch. Detailed information on the dataset construction and hyperparameter configurations for fine-tuning can be found in Appendix \ref{appendix_add1}.

\subsection{Theoretical Analysis of PPL Chunking}
LLMs are designed to learn a distribution $Q$ that approximates the empirical distribution $P$ from sample texts. To quantify the closeness between these two distributions, cross-entropy is typically employed as a metric. Under the discrete scenario, cross-entropy of $Q$ relative to $P$ is formally defined as follows:
\begin{eqnarray}\label{eq:1}
H(P,Q)=\mathrm{E}_{p}[-logQ]=-\sum_{x}P(x)\log{Q(x)} =H(P)+D_{KL}(P||Q)
\end{eqnarray}
where $H(P)$ represents the empirical entropy, and $D_{KL}(P||Q)$ is the Kullback-Leibler (KL) divergence between $Q$ and $P$. The PPL of LLMs, mathematically speaking, is defined as:
\begin{eqnarray}
\text{PPL}(P,Q)=2^{H(P,Q)}
\end{eqnarray}
It is essential to notice that, since $H(p)$ is unoptimizable and bounded as shown in Appendix \ref{appendix1}, what truly impacts the discrepancy in PPL calculations across different LLMs is the KL divergence, which serves as a metric to assess the difference between distributions. The greater the KL divergence is, the larger the disparity between two distributions signifies. Furthermore, high PPL indicates the cognitive hallucination of LLMs towards the real content, and such portions should not be segmented.

On the other hand, \cite{shannon1951prediction} approximates the entropy of any language through a function
\begin{align}\label{eq:3}
G_{K}=&-\sum_{T_{k}}P(T_{k})\log_{2}{P(t_{k}|T_{k-1})}  \nonumber\\
=&-\sum_{T_{k}}P(T_{k})\log_{2}{P(T_{k})}+\sum_{T_{k-1}}P(T_{k-1})\log_{2}{P(T_{k-1})}
\end{align}
where $T_{k}$ represents $k$ consecutive tokens $(t_{1},t_{2}, \dots, t_{k})$ in a text sequence, entropy can then be expressed as
\begin{eqnarray}
H(P)=\lim_{K \to \infty} G_{K}
\end{eqnarray}
Then, based on the proof in Appendix \ref{appendix1} that $G_{K+1}\le G_{K}$ for all $K\ge 1$, we can derive
\begin{eqnarray}\label{eq:2}
G_{1}\ge G_{2}\ge \dots\ge \lim_{K \to \infty} G_{K}=H(P)
\end{eqnarray}
By combining equation \eqref{eq:1} and \eqref{eq:2}, we observe that for large-scale text processing tasks, increasing the context length tends to reduce the cross-entropy or PPL, a phenomenon that reflects the ability of LLMs to make more effective logical inferences and semantic understandings after capturing broader contextual information. Consequently, during PPL Chunking experiments, we maximize the input of longer text sequences to LLMs, anticipating more substantial performance gains.

\section{Experiment}
\subsection{Datasets and Metrics}
We conduct a comprehensive evaluation on five datasets, focusing on both Chinese and English languages, and covering multiple metrics. The LongBench benchmark \cite{bai2023longbench} comprises various datasets, among which we exploit three English datasets and one Chinese dataset, covering both single-hop and multi-hop QA tasks, with evaluations conducted based on F1 and chunking time metrics. The CRUD \cite{lyu2024crud} is a novel benchmark designed for evaluating RAG systems, employing BLEU series, ROUGE-L, and BERTScore metrics for assessment.

\subsection{Baselines}
We primarily compare Meta-Chunking with two types of methods, namely rule-based chunking and dynamic chunking, noting that the latter incorporates both semantic similarity models and LLMs. The original rule-based method simply divides long texts into fixed-length chunks, disregarding sentence boundaries. The Llama\_index method \cite{langchain} offers a more nuanced approach, balancing the maintenance of sentence boundaries while ensuring that token counts in each segment are close to a preset threshold. On the other hand, similarity chunking \cite{xiao2023c} utilizes sentence embedding models to segment text based on semantic similarity, effectively grouping highly related sentences together. Dense X Retrieval \cite{chen2023dense} introduces a new retrieval granularity called propositions, which condenses and segments text by training an information extraction model. Alternatively, LumberChunker \cite{duarte2024lumberchunker} employs LLMs to predict optimal segmentation points within the text. These methods exhibit unique strengths in adapting to the context and structure of texts.

\subsection{Experimental Settings}
\label{sec4_3}
We primarily use Qwen2-0.5B\footnote{\url{https://huggingface.co/Qwen}}, Qwen2-7B\textsuperscript{1} and Baichuan2-7B\footnote{\url{https://huggingface.co/baichuan-inc}} for Meta-Chunking, with Qwen2.5-3B\textsuperscript{1} being employed for fine-tuning. Without additional annotations, all language models used in this paper adopt chat or instruction versions. When chunking, the default parameter configurations of the models are adopted. For evaluation, Qwen2-7B is employed with the following settings: top\_p = 0.9, top\_k = 5, temperature = 0.1, and max\_new\_tokens = 1280. When conducting QA, the system necessitates dense retrievals from the vector database, with top\_k set to 8 for CRUD, and 5 for LongBench. Text chunking is performed on the NVIDIA H800, while model training and evaluation are carried out on the NVIDIA A800. To control variables, we maintain consistent chunk lengths for various chunking methods across each dataset. Detailed experimental setup information can be found in Appendix \ref{appendix3}.

\renewcommand{\arraystretch}{1.1} 
\setlength{\extrarowheight}{1pt} 

\begin{table*}[t]
\centering
\caption{Main experimental results are presented in four QA datasets. The best result is in bold, and the second best result is underlined.}
\label{tab:chunking-method-performance}
\resizebox{\textwidth}{!}{%
\begin{tabular}{lcccccccc}
\toprule
\textbf{Dataset} & \multicolumn{2}{c}{\textbf{2WikiMultihopQA}} & \multicolumn{2}{c}{\textbf{Qasper}} & \multicolumn{2}{c}{\textbf{MultiFieldQA-en}} & \multicolumn{2}{c}{\textbf{MultiFieldQA-zh}} \\
\textbf{Chunking Method} & \textbf{F1} & \textbf{Time} & \textbf{F1} & \textbf{Time} & \textbf{F1} & \textbf{Time} & \textbf{F1} & \textbf{Time} \\
\midrule
\multicolumn{9}{c}{\textit{Baselines with rule-based or similarity-based chunking}} \\ 
\addlinespace[2pt] 
\cdashline{1-9} 
Original& 11.89 & 0.21 & 9.45 & 0.13 & 29.89 & 0.16 & 22.45 & 0.06  \\
Llama\_index& 11.74 & 8.12 & \underline{10.15} & 5.81 & 28.30 & 6.25 & 21.85 & 5.53  \\
Similarity Chunking & 12.00 & 416.45 & 9.93 & 307.05 & 29.19 & 318.41 & 22.39 & 134.80  \\
Dense X Retrieval	& 5.49 & 57633.07 & 8.23 & 39762.54 & 29.72 & 41789.49 & - & -  \\
LumberChunker (Qwen2.5-14B) & 13.28 & 5244.91 & 9.44 & 3777.03 & \underline{33.09} & 3832.04 & \underline{24.35} & 7228.78  \\
\midrule
\multicolumn{9}{c}{\textit{Perplexity Chunking and Margin Sampling Chunking}} \\ 
\addlinespace[2pt] 
\cdashline{1-9}
\rowcolor[rgb]{0.82, 0.89, 0.91} $\text{PPL Chunking (Qwen2-0.5B)}$ & \textbf{13.56} & 140.54 & 9.62 & 65.45 & 31.02 & 79.72 & 23.52 & 64.02  \\
\rowcolor[rgb]{0.82, 0.89, 0.91} $\text{PPL Chunking (Qwen2-7B)}$ & \underline{13.41} & 736.69 & 9.79 & 486.48 & 32.35 & 523.74 & 23.20 & 424.96 \\
\rowcolor[rgb]{0.82, 0.89, 0.91} $\text{PPL Chunking (Baichuan2-7B)}$ & 12.98 & 858.99 & 10.04 & 569.72 & 32.55 & 632.80 & 23.36 & 569.72 \\
\addlinespace[2pt] 
\cdashline{1-9}
\rowcolor[rgb]{0.85, 0.92, 0.99} $\text{MSP Chunking (Qwen2-0.5B)}$ & 12.13 & 1471.31 & \underline{10.15} & 1071.34 & 31.91 & 1103.51 & 23.69 & 888.19 \\
\rowcolor[rgb]{0.85, 0.92, 0.99} $\text{MSP Chunking (Qwen2-7B)}$ & 13.20 & 8781.82 & \textbf{11.37} & 5755.79 & \textbf{33.56} & 6287.31 & \textbf{24.45} & 9746.76 \\
\bottomrule
\end{tabular}%
}
\end{table*}

\section{Results and Analysis}
\subsection{Main Results}
\label{sec5_1}

\textbf{Comparison against Baselines.}\quad We systematically evaluate the performance of five baseline methods, with the results presented in Table \ref{tab:chunking-method-performance}. Compared with traditional rule-based and semantic chunking methods, as well as the state-of-the-art LumberChunker method which leverages Qwen2.5-14B, MSP Chunking exhibits improved and more stable performance. Meanwhile, PPL Chunking demonstrates advantages in balancing performance and processing time. Furthermore, our approach mitigates the current dilemma where text chunking heavily relies on strong instruction-following capabilities. It can even be integrated with a 0.5B SLM without incurring a significant performance decline. This implies that the full potential of SLMs in text chunking tasks has not yet been entirely harnessed. Their notable efficiency and commendable performance warrant further exploration, positioning them as truly practical tools for chunking.

\renewcommand{\arraystretch}{1.1} 
\setlength{\extrarowheight}{1pt} 
\begin{table}[t]
\centering
\caption{Performance of global information compensation mechanism via text chunk rewriting and summary generation based on chunking results.}
\label{tab:Semantic Completion}
\resizebox{\textwidth}{!}{%
\begin{tabular}{lccccccc} 
\toprule
\textbf{Chunking Method}  & \textbf{BLEU-1} & \textbf{BLEU-2} & \textbf{BLEU-3} & \textbf{BLEU-4} & \textbf{BLEU-Avg} & \textbf{ROUGE-L} & \textbf{BERTScore} \\ 
\midrule
Original  & 0.3515 & 0.2788 & 0.2340 & 0.1997 & 0.2548 & 0.4213 & 0.8489 \\
Llama\_index  & 0.3620 & 0.2920 & 0.2480 & 0.2134 & 0.2682 & 0.4326 & 0.8521 \\ 
Similarity Chunking  & 0.3382 & 0.2692 & 0.2257 & 0.1931 & 0.2462 & 0.4131 & 0.8442 \\ 
LumberChunker  & 0.3456 & 0.2781 & 0.2343 & 0.2011 & 0.2542 & 0.4160 & 0.8514 \\ 
\addlinespace[2pt] 
\cdashline{1-8}
Qwen2.5-14B  & 0.3650 & 0.2928 & 0.2469 & 0.2126 & 0.2679 & 0.4351 & 0.8549 \\ 
Qwen2.5-72B & 0.3722 & 0.2985 & 0.2526 & 0.2170 & 0.2743 & 0.4405 & 0.8550  \\ 
\addlinespace[2pt]
\cdashline{1-8}
\rowcolor[rgb]{0.82, 0.89, 0.91} $\text{PPL (Qwen2-7B)}$  & 0.3724 & 0.3012 & 0.2561 & 0.2206 & 0.2774 & 0.4445 & 0.8584 \\ 
\rowcolor[rgb]{0.82, 0.89, 0.91} $\text{PPL (Baichuan2-7B)}$  & 0.3816 & 0.3096 & 0.2631 & 0.2267 & 0.2847 & 0.4520 & 0.8603 \\ 
\addlinespace[2pt] 
\cdashline{1-8}
\rowcolor[rgb]{0.85, 0.92, 0.99} Meta-Chunking & \textbf{0.3924} & \textbf{0.3207} & \textbf{0.2744} & \textbf{0.2378} & \textbf{0.2963} & \textbf{0.4614} & \textbf{0.8657} \\ 
\rowcolor[rgb]{0.85, 0.92, 0.99} \qquad w/o Rewriting & 0.3909 & 0.3179 & 0.2711 & 0.2349 & 0.2934 & 0.4590 & 0.8622 \\ 
\rowcolor[rgb]{0.85, 0.92, 0.99} \qquad w/o Summary & 0.3834 & 0.3122 & 0.2659 & 0.2301 & 0.2876 & 0.4562 & 0.8635 \\ 
\bottomrule
\end{tabular}%
}
\end{table}

\textbf{Effectiveness of Semantic Completion.}\quad To validate the effectiveness of our proposed meta-chunking framework, experiments are conducted on the CRUD benchmark. During the dataset preparation phase, we meticulously structure 20,000 samples for each of the two components through a rigorous processing pipeline. This dataset is then utilized to fine-tune the Qwen2.5-3B model, with the obtained comparative results illustrated in Table \ref{tab:Semantic Completion}. Building upon chunking performance, the meta-chunking framework yield further enhancements to overall system. You can find a more in-depth discussion in Section \ref{sec5_4} and Appendix \ref{appendix_add1}.

\renewcommand{\arraystretch}{1.1} 
\setlength{\extrarowheight}{1pt} 

\begin{wraptable}{r}{0.55\textwidth}
\centering
\caption{Performance comparison of LLMs chunking utilizing two types of Qwen2-7B. \textit{base} represents the basic model, while \textit{inst.} denotes the model fine-tuned with instructions.}
\label{tab:Instruction-Following}
\resizebox{0.55\textwidth}{!}{%
\begin{tabular}{lcccc}
\toprule
\textbf{Dataset} & \textbf{2Wiki} & \textbf{Qasper} & \textbf{MQA-en} & \textbf{MQA-zh}  \\
\textbf{Chunking Method} & \textbf{F1}  & \textbf{F1} & \textbf{F1} & \textbf{F1}  \\
\midrule
\rowcolor[rgb]{0.82, 0.89, 0.91} $\text{MSP Chunking}_{base}$ & 12.02 & 9.86 & 32.24 & 21.55  \\	
\rowcolor[rgb]{0.82, 0.89, 0.91} $\text{MSP Chunking}_{inst.}$ & 13.20 & \textbf{11.37} & \textbf{33.56} & \textbf{24.45}  \\	
\addlinespace[2pt] 
\cdashline{1-5}
\rowcolor[rgb]{0.85, 0.92, 0.99} $\text{PPL Chunking}_{base}$ & \textbf{14.15} & 10.11 & 31.35 & 23.63   \\
\rowcolor[rgb]{0.85, 0.92, 0.99} $\text{PPL Chunking}_{inst.}$ & 13.41 & 9.39 & 32.35 & 23.20 \\
\bottomrule
\end{tabular}%
}
\end{wraptable}

\subsection{Demystifying the Effect of Instruction-Following Capability}
\label{sec5_2}
\begin{figure*}[t]
    \centering
    \includegraphics[width=0.95\textwidth]{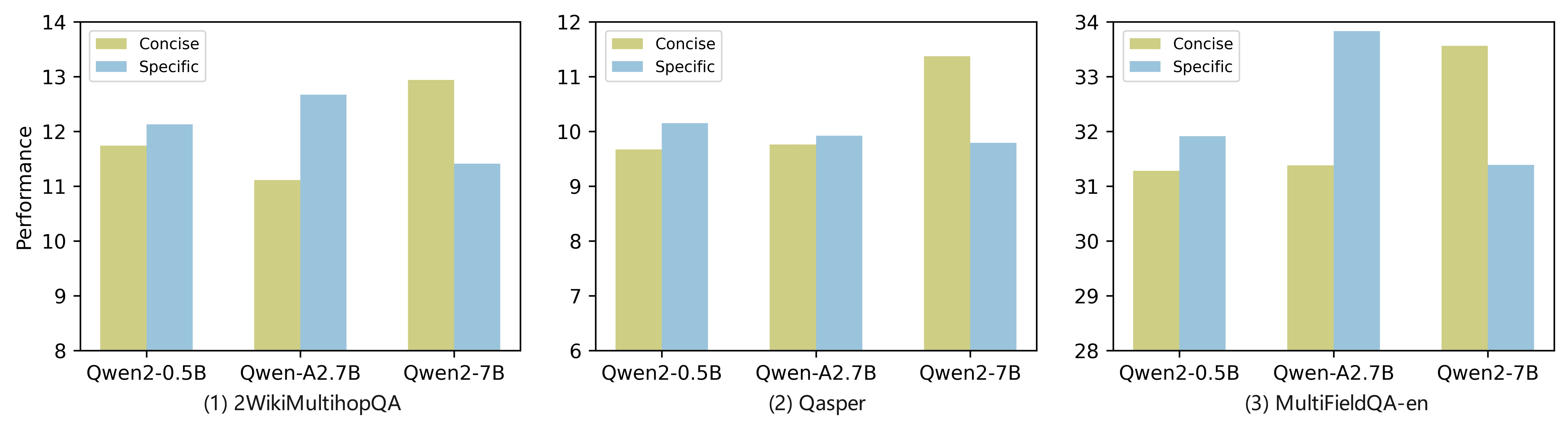}
    \caption{Performance comparison of MSP Chunking using two types of prompts across LLMs of different sizes.}
    \label{fig:MSP_analysis}
\end{figure*}

The experimental results in Section \ref{sec5_1} preliminarily suggest that our method imposes weaker requirements on a model's instruction-following capabilities. However, as pointed out in \cite{he2024does,chang2025influence,srivastava2025revisiting}, prompts influence both the output and reasoning performance of LLMs. Therefore, we conduct a more thorough analysis of the interaction between a model's chunking ability and instructions. As shown in Table \ref{tab:Instruction-Following}, by comparing the base model with the instruction model, we find that the PPL Chunking exhibits greater emphasis on a model's reasoning ability, without imposing stringent requirements on the capability to follow specific instructions. The MSP Chunking, conversely, due to dependency on prompts, emerges a certain degree of need for this ability. Furthermore, we design two types of prompts for MSP Chunking: a regular one and a more precise one, as detailed in Tables \ref{tab:prompt_crud1} and \ref{tab:prompt_crud2}. From Figure \ref{fig:MSP_analysis}, it can be observed that smaller models can benefit from more precise prompts, whereas larger models may experience a decline in performance when subjected to them.

\subsection{Impact of Overlapping Chunking Strategies}
\label{sec5_3}
As demonstrated in Table \ref{tab:multi-hop-overlap}, we investigate the performance of several methods that support overlapping chunks, with their specific implementation details described in Appendix \ref{appendix3}. The dynamic overlap strategy of PPL Chunking assigns sentences located at the minima of the PPL distribution to both the preceding and subsequent chunks, thereby more effectively bridging semantic connections between text chunks. Specifically, apart from a 1\% improvement on the BERTScore, PPL Chunking overlap method achieves performance gains of 2\%-3\% across the remaining metrics.

\renewcommand{\arraystretch}{1.1} 
\setlength{\extrarowheight}{1pt} 
\begin{table*}[h]
\centering
\caption{Performance of different methods on the CRUD benchmark with overlapping chunks.}
\label{tab:multi-hop-overlap}
\resizebox{\textwidth}{!}{%
\begin{tabular}{lcccccccc} 
\toprule
\textbf{Chunking Method} & \textbf{Overlap} & \textbf{BLEU-1} & \textbf{BLEU-2} & \textbf{BLEU-3} & \textbf{BLEU-4} & \textbf{BLEU-Avg} & \textbf{ROUGE-L} & \textbf{BERTScore} \\ 
\midrule
\rowcolor[rgb]{0.82, 0.89, 0.91} Original & Fixed & 0.3330 & 0.2641 & 0.2214 & 0.1881 & 0.2410 & 0.4060 & 0.8425 \\
\rowcolor[rgb]{0.82, 0.89, 0.91} Llama\_index & Dynamic & 0.3326 & 0.2645 & 0.2214 & 0.1890 & 0.2413 & 0.4039 & 0.8439 \\ 
\addlinespace[2pt] 
\cdashline{1-9} 
\rowcolor[rgb]{0.85, 0.92, 0.99} PPL (Qwen2-7B) & Dynamic  & 0.3582 & 0.2898 & 0.2450 & 0.2097 & 0.2657 & 0.4308 & 0.8548 \\ 
\rowcolor[rgb]{0.85, 0.92, 0.99} PPL (Baichuan2-7B) & Dynamic  & 0.3656 & 0.2952 & 0.2497 & 0.2143 & 0.2705 & 0.4393 & 0.8549 \\ 
\bottomrule
\end{tabular}%
}
\end{table*}

\renewcommand{\arraystretch}{1.1} 
\setlength{\extrarowheight}{1pt} 

\begin{wraptable}{r}{0.55\textwidth}
\centering
\caption{Comparison of average similarity scores for different text chunk types across two retrievers.}
\label{tab:Rewrite Analysis}
\resizebox{0.55\textwidth}{!}{%
\begin{tabular}{lcc}
\toprule
\textbf{Metric} & \multicolumn{2}{c}{\textbf{AVGSim@8}}  \\
\textbf{Chunking Method} & \textbf{bge-base-zh-v1.5}  & \textbf{bge-large-zh-v1.5}   \\
\midrule
Original & 0.8053 & 0.8049 \\
Llama\_index & 0.8071 & 0.8032 \\
Similarity Chunking & 0.7751 & 0.7682 \\
LumberChunker & 0.8084 & 0.8034 \\
\addlinespace[2pt] 
\cdashline{1-3}
\rowcolor[rgb]{0.82, 0.89, 0.91} Append Summaries & 0.8246 & 0.8161   \\
\addlinespace[2pt] 
\cdashline{1-3}
\rowcolor[rgb]{0.85, 0.92, 0.99} Rewrite Chunks & \textbf{0.8410} & \textbf{0.8426} \\
\bottomrule
\end{tabular}%
}
\end{wraptable}

\subsection{Rationale for Performance Gains from Text Chunk Rewriting}
\label{sec5_4}
This section aims to elucidate the mechanism by which globally augmented text chunk rewriting enhances system performance. As illustrated in Table \ref{tab:Rewrite Analysis}, we compare different types of text chunks under two distinct semantic retrievers by calculating the average similarity scores for the Top-8 retrieved chunks in response to queries. The results indicate that rewritten chunks exhibit superior alignment with the query intent, thereby facilitating the acquisition of content that is highly relevant to the questions. Furthermore, as depicted by experiments in Figure \ref{fig:ppl_rewrite}, rewritten chunks consistently represent a lower PPL across different LLMs. This phenomenon provides evidence that the global augmentation mechanism enables LLMs to better comprehend retrieved texts by optimizing contextual coherence.

\section{Conclusion}

Addressing issues of logical discontinuity and semantic incompleteness in text chunking, this paper proposes the Meta-Chunking framework, which establishes a systematic solution through the dual constraints of logical perception and information integrity optimization by LLMs. Specifically, we engineer two uncertainty-based adaptive boundary detection algorithms and introduce a dynamic merging strategy to enhance the logical completeness of chunking results. Furthermore, a collaborative information compensation mechanism is developed. It repairs semantic discontinuities caused by segmentation through globally missing-aware rewriting and context-aware summary generation. By autonomously constructing high-quality training datasets and fine-tuning SLMs, this algorithm can be efficiently deployed. Experimental results also corroborate that our framework can achieve higher-quality text chunking while being adaptable to SLMs. We anticipate that our insights will inspire further researches into text chunking, ultimately fostering the development of RAG systems.

\bibliography{refs}
\bibliographystyle{IEEEtran}


\appendix

\section{Theoretical Proof for PPL Chunking}
\label{appendix1}
Firstly, we illustrate the relationship between cross-entropy and two distributions $P$ and $Q$ in another way. Based on sequencing inequality
\begin{eqnarray*}
\sum_{i=1}^{n} a_{i}b_{i}\ge \sum_{i=1}^{n} a_{i}b_{j(i)}\ge \sum_{i=1}^{n} a_{i}b_{n+1-i}
\end{eqnarray*}
where $a_{1}\ge a_{2}\ge \dots\ge a_{n}$, $b_{1}\ge b_{2}\ge \dots\ge b_{n}$ and $(j(1), j(2), \dots , j(n))$ is an arbitrary sorting of $(1,2,\dots,n)$, it can be observed that the sum of products of larger numbers paired together is the maximum, while the sum of products of larger numbers paired with smaller numbers is the minimum. We desire the cross-entropy $H(P,Q)$ to be as small as possible, which means that when $P(x)$ is relatively large, $-\log{Q(x)} $ should be relatively small, thereby resulting in $Q(x)$ also being relatively large. Therefore, a smaller cross-entropy indicates that the prediction is closer to the actual label.

Afterwards, inspired by insights provided in \cite{huyen2019evaluation}, a property of formula \eqref{eq:2} is proved: $G_{K+1}\le G_{K}$ for all $K\ge 1$.
\begin{proof}
\begin{align*}
&~G_{K}-G_{K+1}\\
=&-\sum_{T_{k}}P(T_{k})\log_{a}{P(t_{k}|T_{k-1})} + \sum_{T_{k+1}}P(T_{k+1})\log_{a}{P(t_{k+1}|T_{k})}\\
=&\sum_{T_{k-1}}\left [\sum_{t_{k},t_{k+1}}P(T_{k+1})\log_{a}{P(t_{k+1}|T_{k})}- \sum_{t_{k}}P(T_{k})\log_{a}{P(t_{k}|T_{k-1})} \right ]\\
\ge &\sum_{T_{k-1}}\left [\sum_{t_{k},t_{k+1}}P(T_{k+1})\log_{a}{P(t_{k+1}|T_{k-1})}- \sum_{t_{k}}P(T_{k})\log_{a}{P(t_{k}|T_{k-1})} \right ]\\
= &\sum_{T_{k-1}}\left [\sum_{t_{k},t_{k+1}}P(T_{k-1},t_{k},t_{k+1})\log_{a}{P(t_{k+1}|T_{k-1})}- \sum_{t_{k}}P(T_{k-1},t_{k})\log_{a}{P(t_{k}|T_{k-1})} \right ]\\
= &\sum_{T_{k-1}}\left [\sum_{t_{k+1}}\log_{a}{P(t_{k+1}|T_{k-1})}\sum_{t_{k}}P(T_{k-1},t_{k},t_{k+1})  - \sum_{t_{k}}P(T_{k-1},t_{k})\log_{a}{P(t_{k}|T_{k-1})} \right ]\\
= &\sum_{T_{k-1}}\left [\sum_{t_{k+1}}P(T_{k-1},t_{k+1})\log_{a}{P(t_{k+1}|T_{k-1})}  - \sum_{t_{k}}P(T_{k-1},t_{k})\log_{a}{P(t_{k}|T_{k-1})} \right ]\\
= &0
\end{align*}
The reason for the last equality is that $t_{k+1}$ and $t_{k}$ belong to the same domain. Thus, the proof is complete.
\end{proof}

Eventually, we illustrate bounds of entropy, so as to demonstrate the positive correlation between $H(P,Q)$ and $D_{KL}(P||Q)$ in formula \eqref{eq:1}.
\begin{proof}
Let $P$ be a discrete random variable with a finite range of values denoted by $W := \{w_1,w_{2}, \dots, w_{l}\}$. Set $p_{i} = P\{P = w_{i}\}$ for $i = 1,2, \dots, l$, and assume that $p_{i} > 0$ for all $i \in \{1,2, \dots, l\}$.
According to Lemma 2 in \cite{dragomir1997some}, if 
\begin{eqnarray*}
\gamma :=  \max_{i,j} \frac{\theta_{i}}{\theta_{j}} \le \varphi(\varepsilon ):= 1+\varepsilon\ln{c}+\sqrt{\varepsilon\ln{c}(\varepsilon\ln{c}+2) }  
\end{eqnarray*}
then
\begin{eqnarray*}
0\le \log_{c}{\left (\sum_{k=1}^{l}p_{k}\theta_{k} \right )-\sum_{k=1}^{l}p_{k}\log_{c}{\theta_{k}}}\le \varepsilon  
\end{eqnarray*}
where $\theta_{k}\in (0,+\infty )$, $p_{k}\ge 0$ with $ {\textstyle \sum_{k=1}^{l}} p_{k}=1$ and $c>1$. Given that $\theta_{k}=1/p_{k}$, the aforementioned inequality can be transformed into 
\begin{eqnarray*}
0\le \log_{c}{l} -H_{c}(P)\le \varepsilon 
\end{eqnarray*}
where $\varepsilon>0$ satisfies the following conditions
\begin{eqnarray*}
\max_{i,j} \frac{p_{i}}{p_{j}} \le \varphi(\varepsilon )
\end{eqnarray*}
Furthermore, we can derive bounds for entropy as $\log_{c}{l}-\varepsilon  \le H_{c}(P)\le \log_{c}{l}$. The proof is concluded.
\end{proof}

\begin{figure*}[h]
    \centering
    \includegraphics[width=0.9\textwidth]{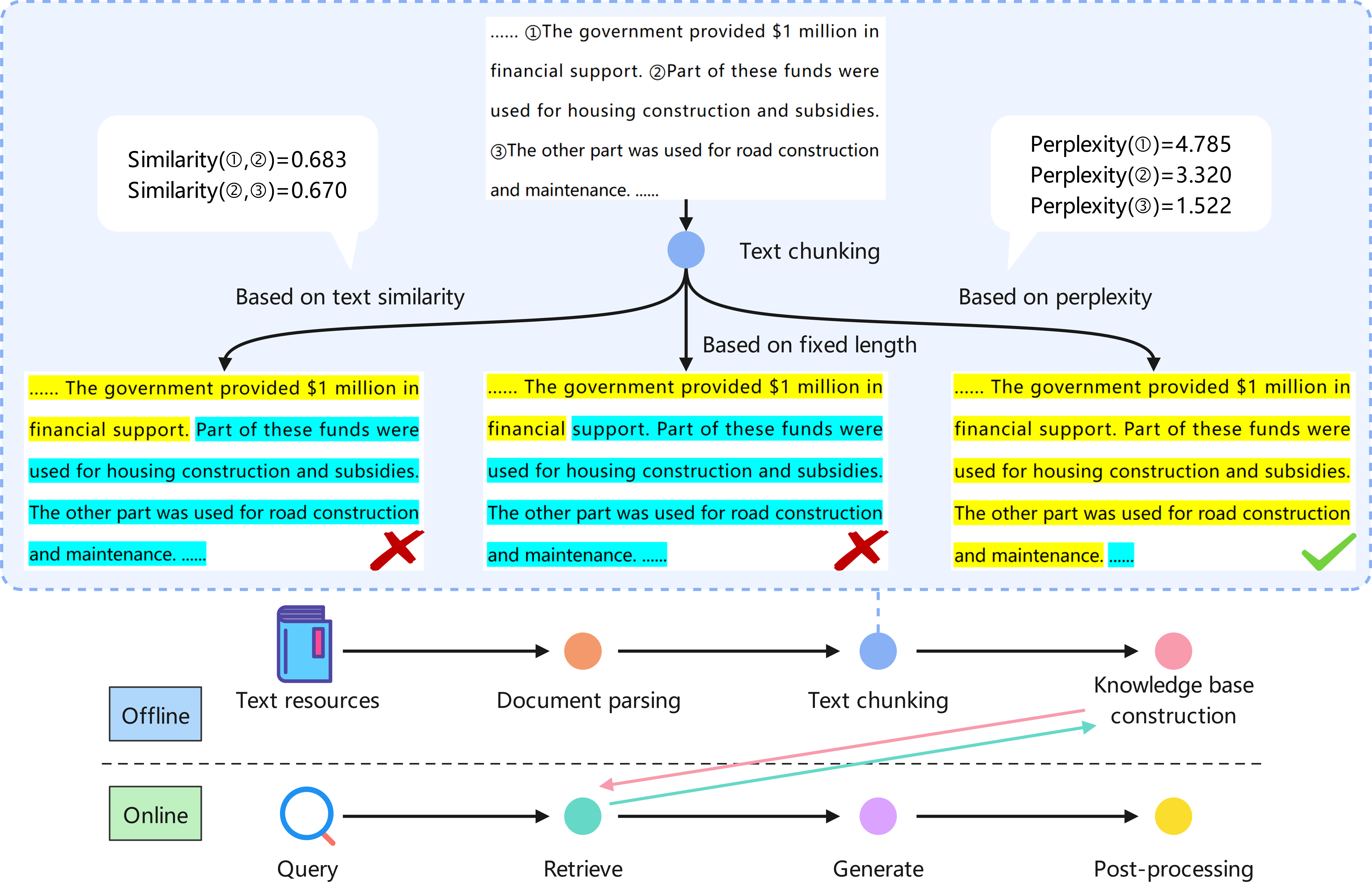}
    \caption{Overview of RAG pipeline, as well as examples based on rules, similarity, and PPL Chunking. The same background color represents being located in the same chunk.}
    \label{fig:rag_pipeline}
\end{figure*}

\section{Design Philosophy of Logical Chunking}
\label{appendix2}
Our approach to text segmentation, centered on logical chunking, distinguishes itself fundamentally from methods primarily reliant on semantic similarity by prioritizing the preservation of complete logical arguments and the integrity of idea expression within each chunk. To ensure logical integrity, our method allows for variable chunk sizes. This dynamic granulation produces chunks that are complete ideational units, thereby preventing logical discontinuities during segmentation, which leads to enhanced document retrieval relevance and improved content clarity.

The key advantage of this logical approach is its ability to recognize and maintain coherence even when constituent sentences exhibit low semantic similarity due to discussing different facets or representations of a core idea. Semantic chunking can falter here, potentially fragmenting a coherent logical argument if the direct semantic overlap between consecutive, logically-linked sentences is not high. In contrast, our method ensures that each meta-chunk is a self-contained logical expression, thereby avoiding breaks in the logical chain. 

\begin{figure*}[h]
    \centering
    \includegraphics[width=0.95\textwidth]{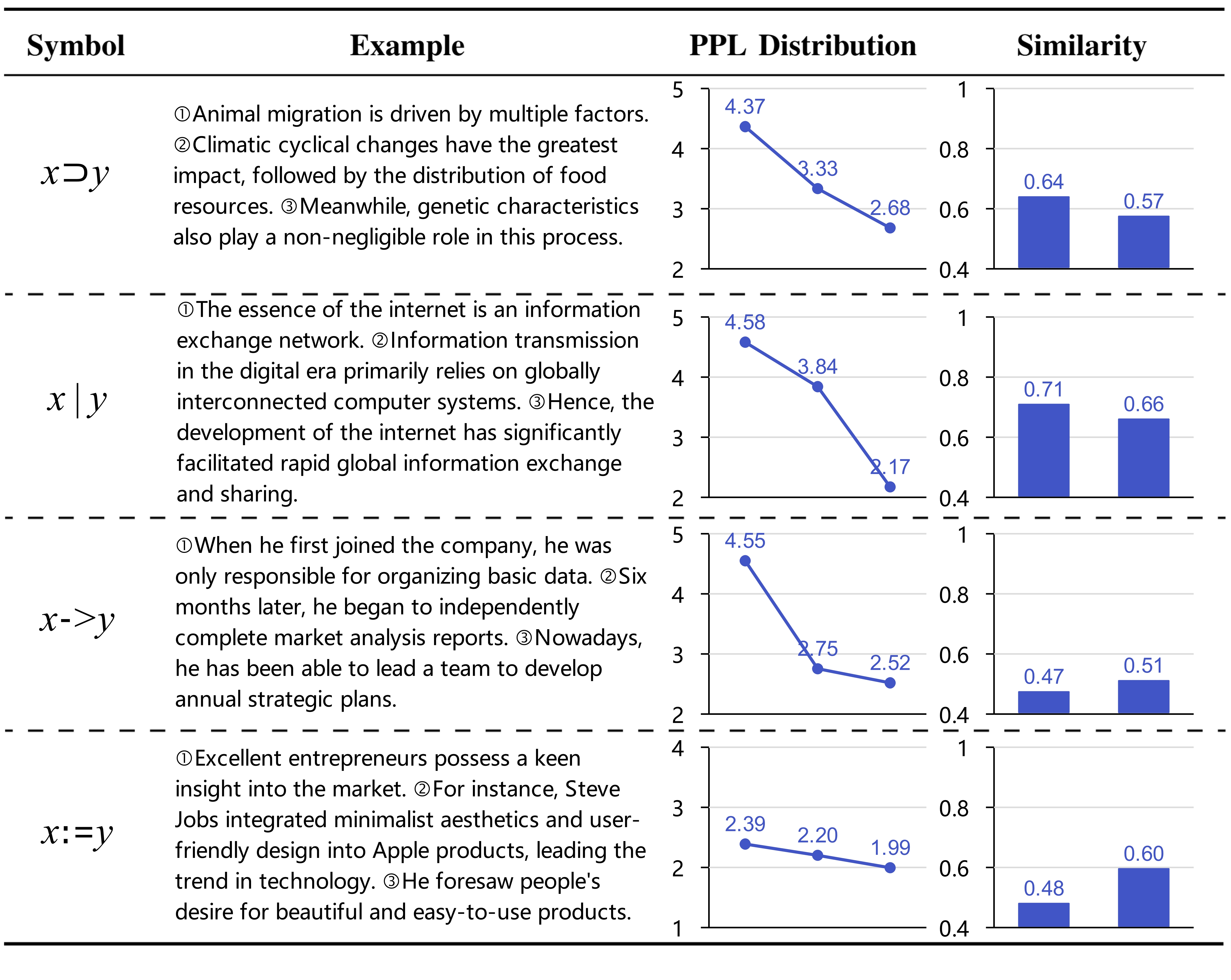}
    \caption{Examples of PPL value variations and semantic similarity for sentences with  different logical relationships, where $x\supset y$, $x|y$, $x->y$, and $x:=y$ refer to general-specific, parallel, sequential, and illustrative relationships, respectively.}
    \label{fig:luoji}
\end{figure*}

As illustrated in Figure \ref{fig:luoji}, among the four scenarios we enumerated, the sentences maintain logical relationships with each other. It can be observed that the PPL distribution based on LLMs exhibits a gradual declining trend, and our chunking method would group these sentences into a single text chunk. However, the semantic similarity between these sentences is relatively low, indicating a high probability of them being separated, which may consequently lead to logical fragmentation.

\section{Detailed Procedure for Semantic Completion}
\label{appendix_add1}
When the original text is segmented into isolated text chunks, each chunk may lose cross-chunk contextual associations, global structural coherence, or implicit logical relationships, leading to the following issues:
\begin{itemize}
    \item \textbf{Incomplete Information}: Critical details are truncated or dispersed across multiple chunks.
    \item \textbf{Semantic Discontinuity}: Logical relationships between chunks are fragmented, impairing the model's comprehension of the overall semantics.
    \item \textbf{Noise Interference}: Irrelevant content is erroneously included within chunks, degrading the accuracy of retrieval and generation tasks.
\end{itemize}

By employing globally enhanced rewriting and summary generation, we can supplement each text chunk with missing global information, bridge semantic gaps, and ultimately elevate the response quality of RAG systems.

During the construction of our training dataset, we initially employ the QwQ-32B\footnote{\url{https://huggingface.co/Qwen/QwQ-32B}} model, leveraging its long-inference mode, to comprehensively identify informational gaps and the requisite supplementary content. Following this, the ERNIE-3.5-128K\footnote{\url{https://console.bce.baidu.com/qianfan/overview}} model is utilized to perform model-based scoring and filtration of this potentially missing information. These refined informational fragments are then fused with the content of the current text chunk, generating a text segment that is both contextually coherent and semantically more complete.

Simultaneously, we leverage the ERNIE-3.5-128K model to generate highly condensed summaries informed by global information. This process aims to enhance the overall contextual awareness of text chunks. Specifically, ERNIE-3.5-128K employs a two-stage strategy: it utilizes document-level global information to generate a supplementary summary for the target text chunk, and concurrently produces a local summary for the text chuk itself. Subsequently, the model meticulously fuses these two types of summaries, ultimately yielding an enhanced summary sentence that clearly articulates the text chunk from a global perspective.

Through this meticulously designed series of processes, we leverage a LLM-driven data distillation pipeline to obtain voluminous and diverse high-quality training samples. At present, we construct 20K data instances for each of the two modules, providing crucial guidance signals for the full fine-tuning of SLMs. This approach enables our framework to uniquely balance high performance with lightweight deployment.

\begin{figure*}[t]
    \centering
    \includegraphics[width=\textwidth]{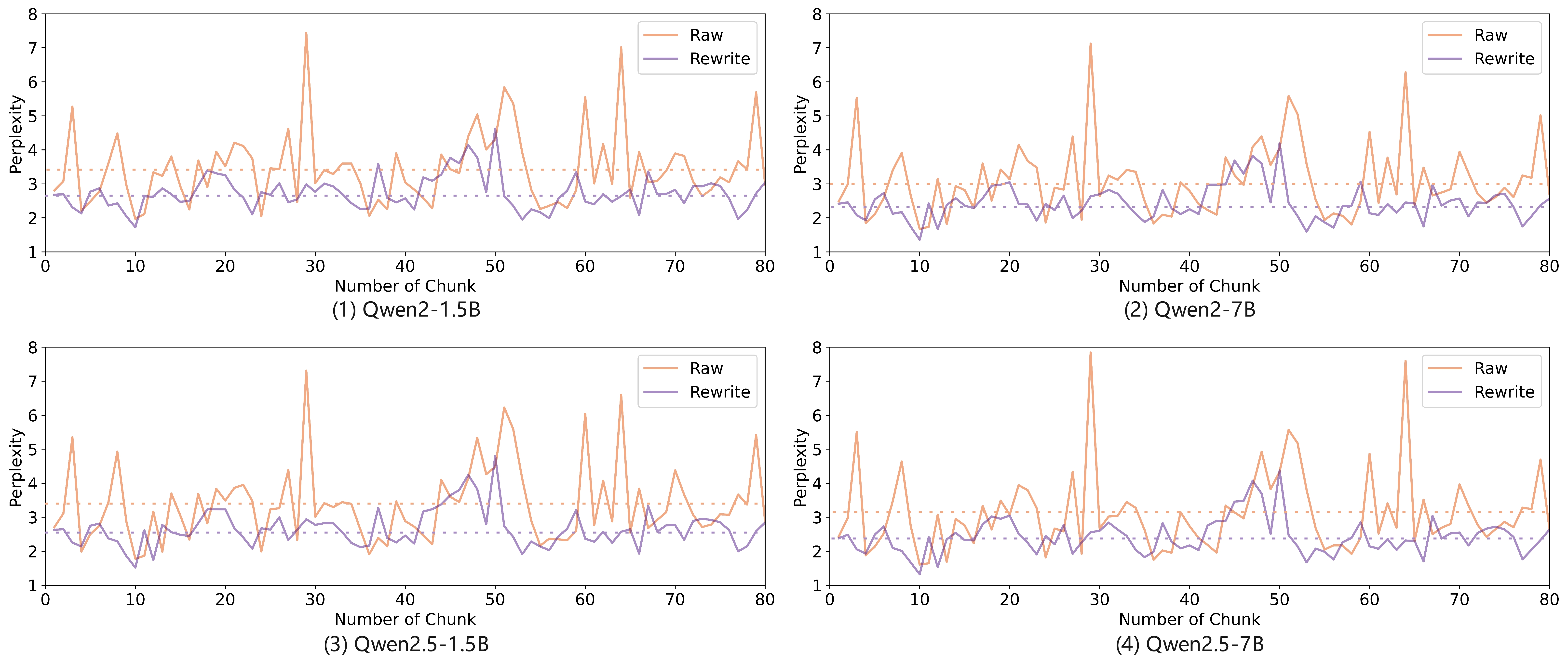}
    \caption{Trends in PPL distribution variations between original and rewritten text chunks across different LLMs.}
    \label{fig:ppl_rewrite}
\end{figure*}

\section{Main Experimental Details}
\label{appendix3}
All language models utilized in this paper employ the chat or instruct versions where multiple versions exist, and are loaded in full precision (Float32). The vector database is constructed using Milvus, where the embedding model for English texts is bge-large-en-v1.5\footnote{\url{https://huggingface.co/BAAI/bge-large-en-v1.5}}, and bge-base-zh-v1.5\footnote{\url{https://huggingface.co/BAAI/bge-base-zh-v1.5}} for Chinese texts. When conducting QA, the system necessitates dense retrievals from the vector database, with top\_k set to 8 for CRUD and 5 for LongBench. In experiments, we utilize a total of five baselines, and their specific configurations are detailed as follows:
\begin{enumerate}[(a)]
    \item \textbf{Rule-based Chunking Methods}
    \begin{itemize}
        \item \textbf{Original}: This method divides long texts into segments of a fixed length, such as two hundred Chinese characters or words, without considering sentence boundaries.
        
        \item \textbf{Llama\_index} \cite{langchain}: This method considers both sentence completeness and token counts during segmentation. It prioritizes maintaining sentence boundaries while ensuring that the number of tokens in each chunk are close to a preset threshold. We use the \texttt{SimpleNodeParser} function from \texttt{Llama\_index}, adjusting the \texttt{chunk\_size} parameter to control segment length. Overlaps are handled by dynamically overlapping segments using the \texttt{chunk\_overlap} parameter, ensuring sentence completeness during segmentation and overlapping.
    \end{itemize}
    
    \item \textbf{Dynamic Chunking Methods}
    \begin{itemize}
        \item \textbf{Similarity Chunking} \cite{xiao2023c}: Utilizes pre-trained sentence embedding models to calculate the cosine similarity between sentences. By setting a similarity threshold, sentences with lower similarity are selected as segmentation points, ensuring that sentences within each chunk are highly semantically related. This method employs the \texttt{SemanticSplitterNodeParser} from \texttt{Llama\_index}. For English texts, we exploit the bge-large-en-v1.5 model, and for Chinese texts, the bge-base-zh-v1.5 model. The size of the text chunks is controlled by adjusting the similarity threshold.
        
        \item \textbf{LumberChunker} \cite{duarte2024lumberchunker}: Leverages the reasoning capabilities of LLMs to predict suitable segmentation points within the text. We utilize Qwen2.5 models with 14B parameters, set to full precision.

        \item \textbf{Dense X Retrieval} \cite{chen2023dense}: Introduces a new retrieval granularity called propositions, which condenses and segments text by training an information extraction model.
    \end{itemize}
\end{enumerate}

In order to control variables during the experiment, we ensure that each dataset have approximately the same chunk size using different chunking methods. Our primary experiments are conducted on the following datasets: 2WikiMultihopQA, Qasper, MultiFieldQA-en, MultiFieldQA-zh, and CRUD, with chunk lengths set to 122, 120, 112, 178, and 178 characters, respectively.

In the Margin Sampling Chunking method, we also use prompt, which mainly consists of two parts: instructions for guiding LLMs to perform chunking and two segmentation schemes. The specific form is shown in Table \ref{tab:prompt_crud1}.
\renewcommand{\arraystretch}{1.2} 
\setlength{\extrarowheight}{1pt} 
\begin{table*}[h]
\centering
\caption{Prompt used in Margin Sampling Chunking.}
\label{tab:prompt_crud1}
\resizebox{\textwidth}{!}{%
\begin{tabular}{p{0.95\textwidth}}
\toprule
\textbf{Chunking Prompt} \\
\hline
This is a text chunking task. You are a text analysis expert. Please choose one of the following two options based on the logical structure and semantic content of the provided sentence:\\
1. Split \textit{sentence1+sentence2} into \textit{sentence1} and \textit{sentence2} two parts;\\
2. Keep \textit{sentence1+sentence2} unsplit in its original form;\\
Please answer 1 or 2.\\
\bottomrule
\end{tabular}%
}
\end{table*}

\renewcommand{\arraystretch}{1.2} 
\setlength{\extrarowheight}{1pt} 
\begin{table*}[h]
\centering
\caption{Prompt with more granular task descriptions for Margin Sampling Chunking.}
\label{tab:prompt_crud2}
\resizebox{\textwidth}{!}{%
\begin{tabular}{p{0.95\textwidth}}
\toprule
\textbf{Chunking Prompt} \\
\hline
This is a text chunking task. You are a text analysis expert. Please group two related paragraphs together and separate unrelated paragraphs based on the logical structure and semantic content of the provided sentences. Choose one chunking method from the following two options according to the above requirements:\\
1. Split \textit{sentence1+sentence2} into \textit{sentence1} and \textit{sentence2} two parts;\\
2. Keep \textit{sentence1+sentence2} unsplit in its original form;\\
Please answer 1 or 2.\\
\bottomrule
\end{tabular}%
}
\end{table*}

As we delve deeper into the influence of text chunking strategies on the performance of complex QA tasks, we further investigate the performance of various chunking strategies when overlapping chunks are employed. The original chunking overlap method uses a fixed number of characters from the end of one chunk to overlap with the start of the next. The Llama\_index overlap approach builds upon this by additionally considering sentence integrity. The PPL Chunking overlap strategy, on the other hand, dynamically assigns sentences represented by minimal points of PPL to both the preceding and subsequent chunks, resulting in dynamic overlap. These approaches generally produce overlap lengths averaging around 50 Chinese characters. Specific experimental results are presented in Section \ref{sec5_3}.

\section{Exploration of Chunking Approach for Performance of Re-ranking}
\label{appendix4}
To explore the impact of chunking strategies on the RAG system, we evaluate the combination of different chunking and re-ranking methods using the MultiHop-RAG benchmark \cite{tang2401multihop}. Initially, a top-10 set of relevant texts is filtered exploiting a dense retriever. We then compare two re-ranking strategies: (1) the BgeRerank method, leveraging the bge-reranker-large model \cite{xiao2023c}, and (2) the PPLRerank method with the Qwen2-1.5B model, utilizing the re-ranking method mentioned in the coarse-grained compression section in \cite{jiang2023longllmlingua}.

\begin{figure*}[h]
    \centering
    \includegraphics[width=1\textwidth]{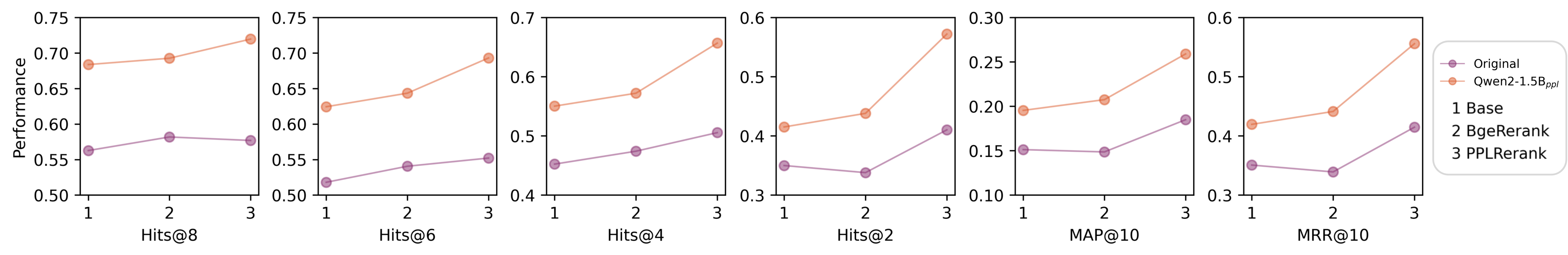}
    \caption{Performance of re-ranking strategies combined with different chunking methods. \textit{ppl} represents direct PPL Chunking, with a threshold of 0.5. The base reveals not utilizing re-ranking strategy.} 
    \label{fig:analysis3}
\end{figure*}

Experimental results in Figure \ref{fig:analysis3} reveal that PPL Chunking and PPLRerank achieve the best overall performance across all metrics. Further analysis demonstrate that, compared to traditional chunking, PPL Chunking not only provide performance gains independently but also significantly enhance the effectiveness of the subsequent re-ranking. Notably, while traditional chunking and re-ranking strategies already deliver performance improvements, PPL Chunking resulted in even greater re-ranking gains. For instance, in the Hits@8 metric, PPLRerank under the original chunking yielded a 1.42\% improvement, whereas PPLRerank under PPL Chunking achieved a 3.59\% improvement. The specific numerical values depicted in the figure can be found in Table \ref{tab:rerank_performance}.

\section{Comparative Analysis of Two PPL Chunking Strategies}
\label{appendix5}
\begin{figure*}[h]
    \centering
    \includegraphics[width=1\textwidth]{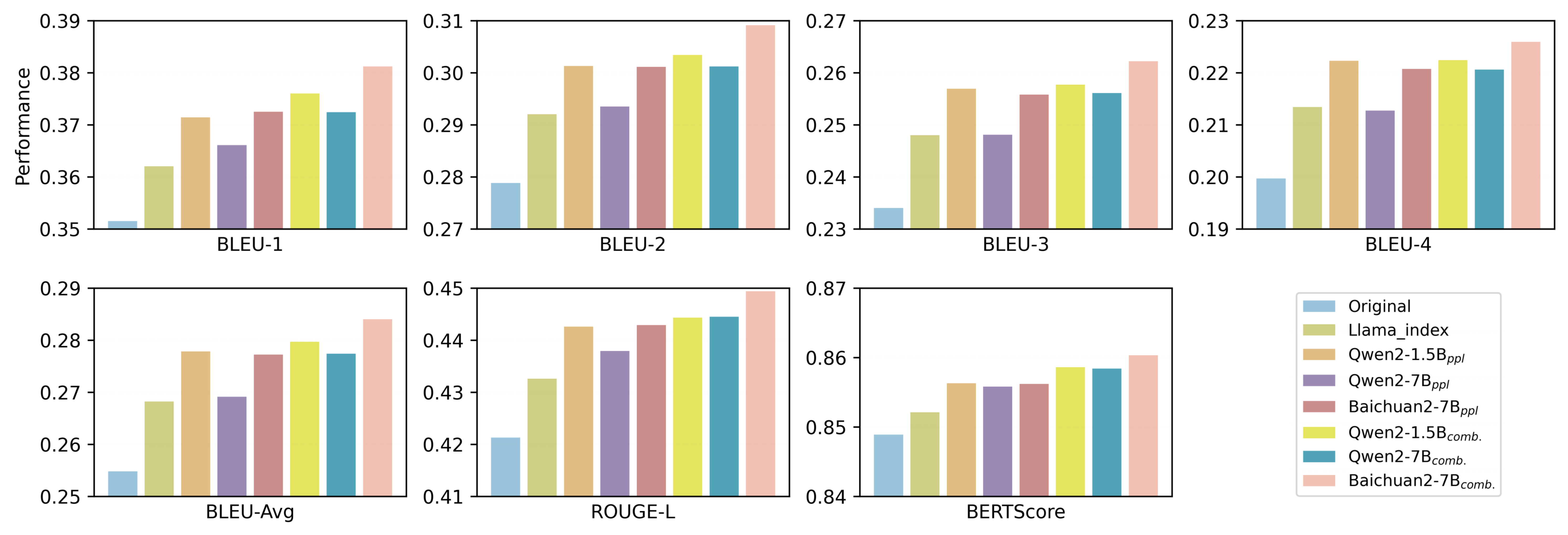}
    \caption{Performance of different methods on the CRUD benchmark. \textit{ppl} represents direct PPL Chunking, with a threshold of 0.5. \textit{comb.} indicates PPL Chunking with dynamic combination, with a threshold of 0 when performing PPL Chunking. }
    \label{fig:analysis4}
\end{figure*}

As shown in Figure \ref{fig:analysis4}, we compare two PPL Chunking strategies: direct PPL Chunking and PPL Chunking with dynamic combination, both of which are effective across the CRUD benchmark. Through experimental analysis, we find that the latter demonstrates superior performance. This is primarily due to direct PPL Chunking, which may result in overly long chunks, whereas the PPL Chunking with dynamic combination method effectively maintains chunk length and logical consistency.

In addition, PPL Chunking achieves significant performance improvements compared to traditional segmentation methods on BLEU series metrics and ROUGE-L. This indicates that our methods enhance the accuracy and fluency of the generated text to the reference text. Furthermore, this experiment reveals the delicate balance between model size and performance. Specifically, the performance of Qwen2-1.5B and Baichuan2-7B under this evaluation framework is closely matched, often surpassing the Qwen2-7B model across multiple metrics. The precise numerical data illustrated in the figure are available in Table \ref{tab:multi-hop11}.

\section{Hyperparameter Selection for PPL Chunking}
\label{appendix6}

\begin{figure*}[h]
    \centering
    \includegraphics[width=1\textwidth]{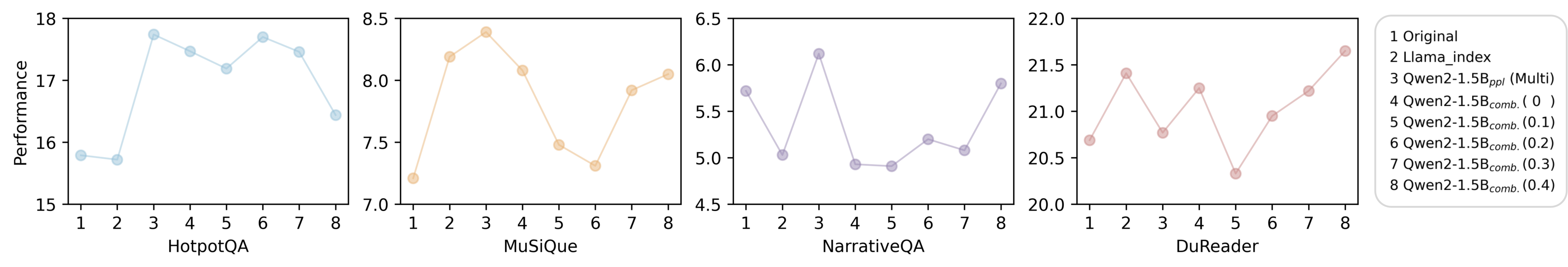}
    \caption{Performance of different methods in four long-text QA datasets of LongBench is evaluated based on F1, F1, F1, and ROUGE-L. \textit{ppl} represents direct PPL Chunking, and \textit{comb.} indicates PPL Chunking with dynamic combination. \textit{Multi} represents threshold values of the parallel method in four datasets, which are 0.5, 0.5, 1.34, and 0.5 respectively, resulting in chunk lengths of 87, 90, 71, and 262 in sequence.}
    \label{fig:analysis2}
\end{figure*}

We conduct an in-depth exploration of chunking in four long-text QA datasets of LongBench, and carry out gradient experiments on the threshold of PPL Chunking, aiming to reveal the intrinsic relationship between PPL distribution and chunking effectiveness. As shown in Figure \ref{fig:analysis2}, when chunk length is small, the direct PPL Chunking brings greater benefits, whereas when the chunk length is longer, PPL Chunking with dynamic combination performs better. In addition, experimental results indicate that the optimal configuration of PPL Chunking relies on the PPL distribution of texts: when the PPL distribution is relatively stable, it is more appropriate to select a lower threshold (such as setting the threshold to 0 in HotpotQA, MuSiQue, and DuReader); whereas when the PPL distribution exhibits large fluctuations, choosing a higher threshold (such as setting the threshold to 0.4 in NarrativeQA) can effectively distinguish paragraphs with different information densities, improving the chunking effect. Therefore, when employing PPL for chunking, it is crucial to comprehensively consider the dual factors of chunk length and text PPL distribution to determine the relatively optimal configuration that maximizes performance.

\section{Collection and Refinement of Training Data}
\label{appendix7}
\subsection{Filtering of Corpora Related to QA Tasks}
In this experiment, we select the QA dataset from the CRUD benchmark. Among them, the single-hop QA dataset consists of questions focused on extracting factual information from a single document. These questions typically require precise retrieval of specific details such as dates, individuals, or events from the provided text. Before the chunking phase, we collect original news articles used in all types of QA tasks in CRUD. Specifically, since CRUD provides evidence context snippets relied on by each QA pair, as well as the original news library where the context snippets are extracted, we can obtain the original news articles containing the context snippets through sentence matching. Taking the two-hop QA as an example, CRUD provides two news snippets, \textit{news1} and \textit{news2}, which are necessary to answer \textit{questions}. We then save the matched original news articles \textit{matched\_news1} and \textit{matched\_news2} that contain \textit{news1} and \textit{news2}. Finally, from the original news library of 80,000 articles, we recall all 10,000 news articles containing context snippets as the initial text for chunking.

\subsection{Dataset Construction for Rewriting and Summary Generation}
To ensure the impartiality and validity of our evaluation, 10K documents obtained through the previously described filtering process are designated as an independent test set. To rigorously prevent data leakage, the dataset used for training the text rewriting and summarization components is entirely sampled from the remaining document corpus, with no overlap with this test set. Specifically, we randomly select 20K long documents from the non-test documents and apply the PPL Chunking via the Baichuan2-7B model for preliminary segmentation. Subsequently, we strategically sample text chunks of varying lengths from each document. Finally, following the data generation pipeline detailed in Section \ref{sec33z}, we prepare the training data for model fine-tuning. The final model performance is evaluated on the reserved independent test set described above.

\section{Corresponding Numerical Values of Images}
\label{appendix8}
In data analysis, the intuitive nature of visual representations facilitates a rapid grasp of the overall landscape. By simultaneously presenting the corresponding numerical values, we provide quantitative foundations for in-depth analysis, enabling a more precise interpretation of experimental results and trends.

\renewcommand{\arraystretch}{1.2} 
\setlength{\extrarowheight}{1pt} 
\begin{table}[h]
\centering
\caption{Performance of re-ranking strategies combined with different chunking methods in the MultiHop-RAG benchmark. \textit{ppl} represents direct PPL Chunking, with a threshold of 0.5.}
\label{tab:rerank_performance}
\resizebox{\textwidth}{!}{%
\begin{tabular}{lcccccc}
\toprule
\textbf{Chunking and Re-ranking} & \textbf{Hits@8} & \textbf{Hits@6} & \textbf{Hits@4} & \textbf{Hits@2} & \textbf{MAP@10} & \textbf{MRR@10} \\ 
\midrule
Original & 0.5627 & 0.5180 & 0.4523 & 0.3499 & 0.1512 & 0.3507 \\ 
Original and BgeRerank & \textbf{0.5818} & 0.5406 & 0.4741 & 0.3379 & 0.1486 & 0.3391 \\ 
Original and PPLRerank & 0.5769 & \textbf{0.5521} & \textbf{0.5055} & \textbf{0.4102} & \textbf{0.1849} & \textbf{0.4147} \\ 
\addlinespace[2pt] 
\cdashline{1-7} 
PPL (Qwen2-1.5B) & 0.6838 & 0.6244 & 0.5503 & 0.4151 & 0.1954 & 0.4195 \\ 
PPL (Qwen2-1.5B) and BgeRerank & 0.6927 & 0.6435 & 0.5721 & 0.4381 & 0.2075 & 0.4413 \\ 
PPL (Qwen2-1.5B) and PPLRerank & \textbf{0.7197} & \textbf{0.6931} & \textbf{0.6568} & \textbf{0.5721} & \textbf{0.2590} & \textbf{0.5558} \\ 
\bottomrule
\end{tabular}%
}
\end{table}

\renewcommand{\arraystretch}{1.2} 
\setlength{\extrarowheight}{1pt} 

\begin{table*}[h]
\centering
\caption{Performance of different methods on the CRUD benchmark. \textit{ppl} represents direct PPL Chunking, with a threshold of 0.5. \textit{comb.} indicates PPL Chunking with dynamic combination, with a threshold of 0 when performing PPL Chunking.}
\label{tab:multi-hop11}
\resizebox{\textwidth}{!}{
\begin{tabular}{lccccccc}
\toprule 						
\textbf{Chunking Method} & \textbf{BLEU-1} & \textbf{BLEU-2} & \textbf{BLEU-3} & \textbf{BLEU-4} & \textbf{BLEU-Avg} & \textbf{ROUGE-L} & \textbf{BERTScore} \\ 
\midrule
Original& 0.3515 & 0.2788 & 0.2340 & 0.1997 & 0.2548 & 0.4213 & 0.8489 \\ 
Llama\_index& 0.3620 & 0.2920 & 0.2480 & 0.2134 & 0.2682 & 0.4326& 0.8521 \\ 
\addlinespace[2pt] 
\cdashline{1-8} 
$\text{Qwen2-1.5B}_{ppl}$ & 0.3714 & 0.3013 & 0.2569 & 0.2223 & 0.2778 & 0.4426 & 0.8563\\ 
$\text{Qwen2-7B}_{ppl}$ & 0.3661 & 0.2935 & 0.2481 & 0.2127 & 0.2691 & 0.4379 & 0.8558 \\ 
$\text{Baichuan2-7B}_{ppl}$ & 0.3725 & 0.3011 & 0.2558 & 0.2207 & 0.2772 & 0.4429 & 0.8562 \\ 
\addlinespace[2pt] 
\cdashline{1-8} 
$\text{Qwen2-1.5B}_{comb.}$ & 0.3760 & 0.3034 & 0.2577 & 0.2224 & 0.2797 & 0.4443 & 0.8586 \\ 
$\text{Qwen2-7B}_{comb.}$ & 0.3724 & 0.3012 & 0.2561 & 0.2206 & 0.2774 & 0.4445  & 0.8584\\ 
$\text{Baichuan2-7B}_{comb.}$ & \textbf{0.3812} & \textbf{0.3091} & \textbf{0.2622} & \textbf{0.2259} & \textbf{0.2840} & \textbf{0.4494} & \textbf{0.8603} \\ 
\bottomrule
\end{tabular}%
}
\end{table*}

\end{document}